\documentclass[11pt]{article}

\usepackage[final]{acl}

\usepackage{times}
\usepackage{latexsym}

\usepackage[T1]{fontenc}

\usepackage[utf8]{inputenc}

\usepackage{microtype}

\usepackage{inconsolata}

\usepackage{graphicx}

\usepackage{amsmath}
\usepackage{amssymb}
\usepackage{amsthm}

\usepackage{booktabs}
\usepackage{multirow}
\usepackage{colortbl}
\usepackage{float}  

\usepackage[ruled,vlined,linesnumbered]{algorithm2e}

\usepackage{enumitem}

\usepackage{listings}
\usepackage{xcolor}

\definecolor{bestblue}{RGB}{0,70,150}     
\definecolor{secondgray}{RGB}{90,90,90}   
\newcommand{\best}[1]{\textbf{\textcolor{bestblue}{#1}}}      
\newcommand{\second}[1]{\underline{\textcolor{secondgray}{#1}}} 

\usepackage{tikz}
\usetikzlibrary{shapes.geometric, arrows.meta, positioning, fit, backgrounds, calc}

\usepackage{hyperref}

\newcommand{\AblSOneHit}{51.6\%}
\newcommand{\AblSOneL}{8.9\%}
\newcommand{\AblSOneHal}{18.4\%}
\newcommand{\AblSOneLat}{0.67s}
\newcommand{\OursAlignedHit}{30.8}
\newcommand{\GptDirAlignedHit}{30.9}
\newcommand{\GptRagAlignedHit}{30.9}
\newcommand{\QwenDirAlignedHit}{26.0}
\newcommand{\QwenRagAlignedHit}{26.0}

\newcommand{\BmAlignedHit}{23.4}
\newcommand{\OursRandHit}{59.3}      
\newcommand{\BmHit}{23.5}       
\newcommand{\GptDirHit}{34.4}   
\newcommand{\GptRagHit}{34.4}   
\newcommand{\QwenDirHit}{28.1} 
\newcommand{\QwenRagHit}{28.1}  

\title{Intent-Driven Semantic ID Generation for Grounded\\ Conversational News Recommendation}

\author{
  \textbf{Hongyang Su\textsuperscript{2}},
  \textbf{Beibei Kong\textsuperscript{2}},
  \textbf{Lei Cheng\textsuperscript{2}},\\
  \textbf{Chengxiang Zhuo\textsuperscript{2}},
  \textbf{Zang Li\textsuperscript{2}},
  \textbf{Chenyun Yu\textsuperscript{1}\thanks{Corresponding author.}}
\\
\\
  \textsuperscript{1}Shenzhen Campus of Sun Yat-sen University, Shenzhen, China
\\
  \textsuperscript{2}Platform and Content Group, Tencent, Shenzhen, China
\\
  \texttt{\{supersusu,echokong,raycheng,felixzhuo,gavinzli\}@tencent.com}
\\
  \texttt{yuchy35@mail.sysu.edu.cn}
}

\begin{document}
\maketitle
\begin{abstract}
Conversational news recommendation requires grounding each suggestion in a rapidly evolving article corpus while addressing implicit user intents that lack explicit retrievable keywords.
To characterize this scenario, we identify 6 intent types from production dialogues: five are implicit and pose fundamental challenges to standard RAG pipelines, forming a critical \textit{retrieve-first} bottleneck.
To address these issues, we introduce \textbf{intent-driven Semantic ID (SID) generation} under a \textbf{Generate-then-Match} paradigm. With two-stage training that consists of multi-task SID alignment and GPT-4 Chain-of-Thought distillation, an LLM maps diverse intents to hierarchical SID prefixes, which are then fuzzy-matched to the current news pool to guarantee fully grounded recommendations. \textbf{Profile-Aware Dual-Signal Reasoning (PADR)} further enables cold-start users to obtain valid recommendations using only profiles.
On a mainstream Chinese news platform, our 7B model achieves \textbf{0\% hallucination} and \textbf{12.4\% L1 match} in the 152K open-generation SID space ($4{\times}$ random baseline). It matches GPT-4+Hybrid RAG on L1 while surpassing it on finer-grained metrics (L2 $2{\times}$, Category $+$1.2pp) at ${\sim}$100$\times$ lower cost. Cold-start users, where existing baselines score 0\%, achieve \textbf{18.0\% L1} ($6{\times}$ random), the highest among all user groups.
\end{abstract}

\section{Introduction}
\label{sec:introduction}

On a mainstream Chinese news platform, we develop a \textbf{chat-based news assistant} that provides personalized recommendations via natural-language dialogue. Unlike stable-catalog recommendation domains (e.g., movies, products), this scenario poses a critical challenge: \textbf{most articles expire within 24 hours} (observed from the fraction of daily impressions from articles published in the past 24h), leading to pervasive cold-start issues for both items and users. Moreover, even returning users become ``cold'' when encountering fresh content, with 20--30\% of users having fewer than 10 interactions. When users request ``recommend news'' or ``something different'', no retrievable keywords are available, yet such \textit{implicit intents} dominate real-world conversations.

We analyze real-world dialogues on our news platform and identify \textbf{6 conversational intent types}. Only explicit keyword queries can be effectively handled by standard RAG, while the remaining 5 categories (i.e., implicit needs, continuation, diversity, feedback, and cold-start) demand sophisticated query construction from user profiles and histories, greatly increasing pipeline complexity.
Even embedding-based retrieval~\cite{lewis2020retrieval,gao2023retrieval} yields limited gains on these intents, as the query lacks discriminative intent-specific content when users say ``recommend something'' or ``something different''.
While generative SID recommendation~\cite{rajput2024recommender,zhou2025onerec} encodes items as learnable hierarchical tokens, it targets \textit{single-turn} next-item prediction and relies on rich behavioral sequences. Extending SIDs to conversational recommendation requires solving two key open problems: (1) generating SID prefixes from diverse dialogue intents (not just click sequences), and (2) reasoning for cold-start users who lack the behavioral history relied upon by SID-based models.

\textbf{Our Approach: Intent-Driven SID Generation.}
We propose \textbf{Generate-then-Match}, a novel paradigm that departs from conventional retrieve-first pipelines. Rather than constructing explicit queries for candidate retrieval, our LLM directly generates SIDs (hierarchical tokens encoding news semantics via RQ-VAE~\cite{rajput2024recommender}) from user intent, profiles, and dialogue context. The generated SID prefixes are then fuzzy-matched to the current news pool, ensuring each recommendation corresponds to a real article \textit{by construction}. To handle the full spectrum of user states, we introduce \textbf{Profile-Aware Dual-Signal Reasoning (PADR)}, which adaptively routes through warm, hybrid, or cold reasoning paths, enabling even zero-history users to receive meaningful recommendations from their profiles alone.
Our contributions are summarized as follows:

\begin{itemize}[nosep,leftmargin=*]
    \item \textbf{Intent-Driven Generate-then-Match}: We propose a two-stage training paradigm that conducts multi-task SID alignment and then applies GPT-4 Chain-of-Thought distillation. It encourages an LLM to generate grounded SID prefixes over all 6 intent types and fully discards conventional retrieve-first pipelines.
    
    \item \textbf{Profile-Aware Dual-Signal Reasoning (PADR)}: We design an adaptive reasoning framework that dynamically selects warm, hybrid, or cold paths based on accessible user signals. It effectively supports cold-start users where existing baselines often fail and yields competitive recommendation performance.
    
    \item \textbf{Industrial-Scale Validation}: Evaluations on a 163K-article test pool show 12.4\% L1 score, outperforming the production baseline of 11.6\% with $p<0.05$. Our solution achieves 0\% hallucination and sub-100ms end-to-end latency.
\end{itemize}

\section{Related Work}
\label{sec:related_work}

\paragraph{Conversational Recommendation Systems.}
CRS elicit user preferences through multi-turn dialogue~\cite{jannach2021survey,gao2021advances}.
From ReDial~\cite{li2018redial} to knowledge-enhanced~\cite{chen2019kbrd,ma2021crwalker}, unified~\cite{wang2022unicrs,zhou2022c2crs}, and LLM-based methods~\cite{friedman2023leveraging,he2023large,gao2023chatrec,wang2023iEvaLM}, most work assumes \textit{stable catalogs} with explicit preference mentions. Our setting differs fundamentally: \textit{ephemeral} news (most articles cycle out within 24h), 5/6 implicit intents lacking retrievable keywords, and cold-start users (see Appendix~\ref{sec:crs-comparison} for a detailed comparison).

\paragraph{Generative Recommendation and Semantic IDs.}
Semantic IDs represent items as learnable hierarchical tokens~\cite{rajput2024recommender,tay2022transformer,zheng2024adapting}. TIGER~\cite{rajput2024recommender} pioneered this approach but assumes stable catalogs, limiting its effectiveness in rapidly refreshing domains like news. The OneRec series~\cite{zhou2025onerec} demonstrates industrial-scale SID deployment, while OneRec-Think~\cite{liu2025onerec} and ThinkRec~\cite{yu2025thinkrec} extend SID-based recommendation with explicit reasoning. DiscRec~\cite{liu2025discrec} disentangles semantic and collaborative signals. For a broader overview, see~\cite{li2024generative}. These methods target \textit{single-turn} next-item prediction and rely on behavioral sequences; our work differs in targeting \textit{conversational} recommendation with 6 intent types, enabling cold-start via PADR, and structurally ensuring 0\% hallucination via Generate-then-Match. We directly compare with TIGER and OneRec in Table~\ref{tab:coldstart_gen}.

\paragraph{Grounding and Hallucination in LLM Recommendations.}
Hallucination remains critical~\cite{zhang2023hallucination,ji2023survey,huang2023hallucination}. RAG~\cite{lewis2020retrieval,gao2023retrieval} mitigates it but fails for implicit queries; constrained decoding~\cite{hokamp2017lexically} limits expressiveness; graph-based retrieval~\cite{qiu2025graph} still requires successful retrieval. Our Generate-then-Match addresses this structurally: the matching function guarantees every output exists in the current news pool (\S\ref{sec:generate-then-match}).

\section{Method}
\label{sec:method}

\begin{figure*}[t]
\centering
\includegraphics[width=\textwidth]{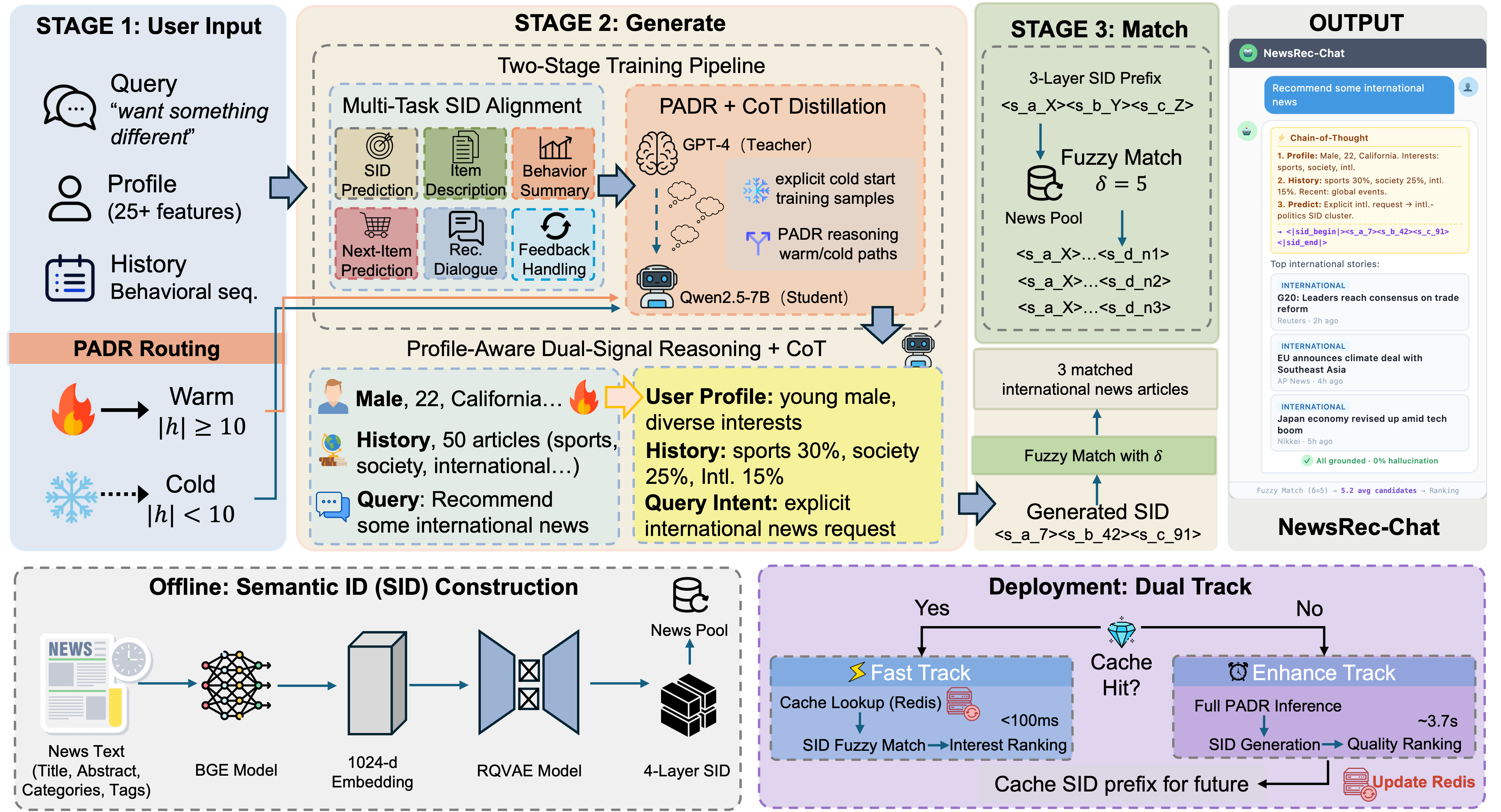}
\caption{Overview of the Generate-then-Match framework. \textbf{Left}: PADR routes user context through warm/cold/hybrid reasoning paths. \textbf{Center}: The two-stage trained LLM (Stage~1: SID alignment, Stage~2: CoT distillation) generates 3-layer SID prefixes from intent and context. \textbf{Right}: Fuzzy matching grounds prefixes to the current news pool. Bottom: offline SID construction (RQ-VAE) and Dual-Track deployment with SID-prefix caching.}
\label{fig:architecture}
\end{figure*}

Figure~\ref{fig:architecture} overviews our Generate-then-Match framework. PADR (\S\ref{sec:padr}) routes user context through warm, hybrid, or cold reasoning paths. The fine-tuned LLM then generates 3-layer SID prefixes encoding the inferred intent, and fuzzy matching ($\delta{=}5$) grounds them to the current news pool (\S\ref{sec:sid-fuzzy}). A Dual-Track architecture caches pool-agnostic prefixes for sub-100ms latency (\S\ref{sec:dual-track}).

\subsection{Generate-then-Match: Replacing Retrieve-First}
\label{sec:generate-then-match}

Traditional RAG follows a \textit{retrieve-then-generate} pipeline: the user query $q$ serves as a retrieval key to fetch candidates. This works when $q$ contains explicit keywords, but 5 of our 6 intent types are \textit{implicit} (Table~\ref{tab:intents}), carrying no retrievable keywords. We reverse this pipeline with a \textit{generate-first} strategy:

\begin{table}[t]
\centering
\footnotesize
\resizebox{\columnwidth}{!}{%
\begin{tabular}{@{}llc@{}}
\toprule
\textbf{Intent Type} & \textbf{Example Query} & \textbf{Explicit?} \\
\midrule
Candidate Selection & ``recommend tech news'' & \checkmark \\
Next-Item Continuation & ``what else?'' & $\times$ \\
Diversity Seeking & ``something different'' & $\times$ \\
Feedback Adjustment & ``not sports'' & $\times$ \\
Cold-Start (PADR) & ``recommend news'' (sparse history) & $\times$ \\
Pure Cold-Start & ``recommend news'' (new user) & $\times$ \\
\bottomrule
\end{tabular}}
\caption{Six conversational intents mapped to evaluation tasks. ``Explicit?'' indicates whether the query contains retrievable keywords. 5/6 intents are implicit, posing a structural challenge for retrieval-first pipelines.}
\label{tab:intents}
\end{table}
An inter-annotator agreement study (Fleiss' $\kappa{=}0.81$, 6-way; Appendix~\ref{sec:iaa-study}) confirms the reliability of this taxonomy.
\begin{equation}
\begin{aligned}
\text{RAG}: & \quad \mathcal{R}(q) \rightarrow \textit{LLM}(\mathcal{R}(q), q) \\
\text{Ours}: & \quad \textit{LLM}(u, h, q) \rightarrow \textit{Match}(\textit{SID}, \mathcal{P})
\end{aligned}
\label{eq:paradigm}
\end{equation}
where $u$ is user profile, $h$ is behavioral history, $q$ is query, $\textit{SID}$ is the generated Semantic ID prefix, and $\mathcal{P}$ is the current news pool. Rather than retrieving first, the LLM \textit{internalizes} user context $(u, h, q)$ and directly generates a structured SID that encodes the inferred intent. The matching function then provides a \textit{pool-existence guarantee}:
\begin{equation}
\textit{Match}(\textit{SID}, \mathcal{P}) \subseteq \mathcal{P}
\label{eq:grounding}
\end{equation}
That is, every recommendation is restricted to $\mathcal{P}$ by construction, eliminating hallucination at the architectural level. This paradigm shift introduces two new challenges: (1)~the model must learn to generate \textit{semantically meaningful} SID prefixes from diverse intents, addressed by two-stage training and PADR (\S\ref{sec:padr}); (2)~generated prefixes must be grounded to a \textit{dynamically refreshed} news pool, addressed by fuzzy matching (\S\ref{sec:sid-fuzzy}).

\subsection{Semantic ID Structure and Fuzzy Matching}
\label{sec:sid-fuzzy}

\paragraph{4-Layer Hierarchical SID.}
We adopt the production SID system already deployed on the platform, where each article is encoded as $\text{SID} = \langle s_1, s_2, s_3, s_4 \rangle$~\cite{rajput2024recommender,tay2022transformer}. SIDs are generated by RQ-VAE over news content embeddings: each layer is a purely \textit{semantic} quantization code capturing progressively finer content similarity, ranging from broad semantic regions ($s_1$) through mid-level groups ($s_2$) and fine-grained clusters ($s_3$) to near-unique article identifiers ($s_4$). This hierarchy enables generation and matching at different granularities (codebook details in Appendix~\ref{sec:training-config}).

\paragraph{3-Layer Prefix Generation.}
The model generates \textbf{3-layer prefixes} $P = (s_1, s_2, s_3)$ rather than full 4-layer SIDs. The first three layers capture \textit{intent-level} semantics: articles sharing $(s_1, s_2, s_3)$ are topically interchangeable. The fourth layer $s_4$ serves as a near-unique article discriminator that shifts daily as articles enter and leave the pool; predicting it would couple the model to a specific day's inventory. Generating up to $s_3$ thus provides the right granularity: specific enough to reflect user intent, yet stable across pool refreshes.

\paragraph{Fuzzy Matching Algorithm.}
Generated prefixes are matched against the pool with tolerance $\delta$:
\begin{equation}
\small
\textit{Match}(P, \mathcal{P}) \!=\! \{n \!\in\! \mathcal{P} : s_1'\!=\!s_1, s_2'\!=\!s_2, |s_3'\!-\!s_3| \!\leq\! \delta\}
\label{eq:fuzzy}
\end{equation}
where $(s_1', s_2', s_3', s_4')$ is the SID of news $n$. Layers 1--2 are matched strictly to preserve semantic coherence, as adjacent $s_1$ or $s_2$ codes may encode unrelated content regions. Layer $s_3$ admits tolerance $\delta$ because neighboring codes within the same $(s_1, s_2)$ group share fine-grained topical similarity. We set $\delta{=}5$ via grid search over $\{1, 3, 5, 7, 10\}$ on a held-out set (Appendix~\ref{sec:fuzzy-rationale}). Matched candidates are ranked by $1 - |s_3' - s_3|/(\delta+1)$, yielding a small candidate set (mean 5.2, median 3.0 articles; Appendix~\ref{sec:partial-match-analysis}).

\subsection{Profile-Aware Dual-Signal Reasoning (PADR)}
\label{sec:padr}

Cold-start users (20--30\% of active users) lack behavioral history. PADR adaptively leverages two complementary signal types based on user state.

\paragraph{Dual-Signal Context Construction.}
Let $u$ denote a user with profile $\mathbf{p}_u$ and behavioral history $\mathbf{h}_u = [n_1, \ldots, n_t]$ (clicked news). The profile $\mathbf{p}_u$ aggregates 25{+} features covering demographics, preferences, and behavioral patterns (Appendix~\ref{sec:profile-features-analysis}). The LLM input context adaptively includes or omits behavioral history:
\begin{equation}
\mathbf{x}_u \!=\!
\begin{cases}
[\mathbf{p}_u; \mathbf{h}_u; q] & |\mathbf{h}_u| \!\geq\! \tau \;\text{(warm)} \\
[\mathbf{p}_u; \mathbf{h}_u; q; \text{``sparse''}] & 0 \!<\! |\mathbf{h}_u| \!<\! \tau \;\text{(hybrid)} \\
[\mathbf{p}_u; q; \text{``no history''}] & |\mathbf{h}_u| \!=\! 0 \;\text{(cold)}
\end{cases}
\label{eq:adaptive}
\end{equation}
where $\tau{=}10$ is the history sufficiency threshold (Appendix~\ref{sec:threshold-sensitivity}). An explicit availability indicator guides the model toward the appropriate reasoning strategy.

\paragraph{Learned Reasoning Paths.}
Unlike rule-based cold-start strategies that hard-code fallback logic, PADR reasoning paths are \textit{learned} through CoT supervision: the model autonomously develops differentiated strategies for each user state. The \textit{Warm Path} ($|\mathbf{h}_u| \geq \tau$) correlates behavioral patterns with profile features; the \textit{Cold Path} ($|\mathbf{h}_u| = 0$) infers interests from demographics alone (achieving 18.0\% L1 accuracy); the \textit{Hybrid Path} cross-references sparse clicks with profiles. No explicit routing module is needed, as the availability indicator in Eq.~\ref{eq:adaptive} suffices for path selection.

\paragraph{Two-Stage Training Pipeline.}
\textit{Stage 1: Multi-Task SID Alignment} builds a grounded SID vocabulary via 6 objectives in 3 groups: \textbf{SID Understanding} (bidirectional content$\leftrightarrow$SID mapping), \textbf{User Modeling} (behavior summarization and next-item prediction), and \textbf{Recommendation Dialogue} (multi-turn recommendation integrating all capabilities; task ratios in Appendix~\ref{sec:task-details}).

\textit{Stage 2: PADR CoT Distillation} then teaches the model to \textit{reason} about which SIDs to generate. GPT-4 produces gold CoT traces for each (input, target SID) pair~\cite{li2023symbolic,magister2023teaching}, which are distilled into the smaller LLM. Three design choices are critical: (1)~\textbf{${\sim}$31\% cold-start samples} to ensure profile-only reasoning; (2)~\textbf{intent-differentiated CoT}, where each intent type receives a distinct reasoning structure (e.g., demographic$\rightarrow$interest for cold-start, preference-shift analysis for feedback); and (3)~\textbf{controlled CoT length} (150--300 chars) to prevent reasoning degradation at inference (\S\ref{sec:cold_start}). Human evaluation of 200 sampled CoTs confirms high quality (mean 4.1/5; Appendix~\ref{sec:cot-quality}).

\subsection{Dual-Track Architecture for Industrial Deployment}
\label{sec:dual-track}

Full PADR inference is too slow for real-time conversation. We design a dual-track architecture to decouple latency from reasoning quality. The \textbf{Fast Track} looks up cached SID prefixes, applies fuzzy matching against the current pool, and ranks candidates; on cache miss, a profile-based fallback provides immediate results. The \textbf{Enhance Track} runs asynchronously, performing full PADR inference and updating the cache.

Two design choices are critical. First, cached SID prefixes are \textit{pool-agnostic}, encoding user intent rather than specific article identifiers, so they remain valid across daily pool refreshes. Second, the Enhance Track proactively generates SID prefixes using \textit{preference-based preset queries} derived from each user's profile, ensuring recommendation timeliness even before the user initiates a conversation (Appendix~\ref{sec:dual-track-details}).

\section{Experiments}
\label{sec:experiments}

\subsection{Setup}

\paragraph{Dataset.}
We use data from a major Chinese news platform (Table~\ref{tab:data}). A \textit{temporal split} ensures zero item overlap between training and test: training uses news before a cutoff date; test uses only news published after. Our \textbf{primary evaluation} is open generation (76\% of test data), where the model generates SID prefixes in the full 152K space. Candidate selection (24\%) uses fixed candidates per sample to isolate discriminative ability (Appendix~\ref{sec:cand-sel-rationale}).

\begin{table}[t]
\centering
\small
\begin{tabular}{@{}llr@{}}
\toprule
\textbf{Data} & \textbf{Description} & \textbf{Scale} \\
\midrule
News Pool & Articles with SID & 163,560 \\
User Profiles & Rich interest profiles & 50,000 \\
Stage 1 Training & SID alignment & 483,374 \\
Stage 2 Training & PADR reasoning & 48,249 \\
Test Samples & 6 task types & 9,982 \\
\bottomrule
\end{tabular}
\caption{Dataset statistics.}
\label{tab:data}
\end{table}

\paragraph{Baselines.}
We compare against five categories of methods.
\textit{Non-Personalized}: Random, Popular, and Hist-Pop (Appendix~\ref{sec:user-history-baseline}).
\textit{LLM-Based}: GPT-4 Direct and Qwen-7B Direct (no retrieval); GPT-4+RAG and Qwen-7B+RAG with BM25~\cite{robertson2009bm25}, BGE-large-zh Dense~\cite{xiao2023bge}, and Hybrid retrieval (Appendix~\ref{sec:dense-retrieval-details}).
\textit{Generative Recommendation}: TIGER~\cite{rajput2024recommender} and OneRec-7B~\cite{zhou2025onerec} (same Qwen2.5-7B backbone with LoRA, constrained SID decoding; Table~\ref{tab:coldstart_gen}).
\textit{Sequential Recommenders}: SASRec~\cite{kang2018sasrec} and BERT4Rec~\cite{sun2019bert4rec} (Appendix~\ref{sec:seq-comparison}).
Our model is fine-tuned from \textbf{Qwen2.5-7B-Instruct}~\cite{qwen2.5} on 4$\times$H20-96G GPUs (Appendix~\ref{sec:training-config}).

\paragraph{Metrics.}
\textit{Candidate Selection}: \textbf{Hit@1} under \textbf{Rand} (4 random negatives) and \textbf{Align} (2 same-category + 2 random) settings.
\textit{Open Generation} (152K SID space): \textbf{L1 Match} (32 categories), \textbf{L2 Match} (2,048 topics), \textbf{Category Match} (Cat., editorial taxonomy alignment).
\textit{Grounding}: \textbf{Hallucination Rate} (Hal., fraction of generated SIDs absent from the current pool).
\textit{System}: Average and P95 latency.

\subsection{Main Results}

Table~\ref{tab:main_results} presents results on the 9,982-sample test set. Generative recommendation baselines (TIGER, OneRec-7B) and sequential recommenders (SASRec, BERT4Rec) are compared in Table~\ref{tab:coldstart_gen} and Appendix~\ref{sec:seq-comparison}, respectively.

\begin{table}[t]
\centering
\small
\setlength{\tabcolsep}{2pt}
\begin{tabular}{@{}ll cc ccc c@{}}
\toprule
\multicolumn{2}{l}{\multirow{2}{*}{\textbf{Method}}} & \multicolumn{2}{c}{\textbf{Cand.\ Hit@1}} & \multicolumn{3}{c}{\textbf{Open Gen.}} & \multirow{2}{*}{\textbf{Hal.}} \\
\cmidrule(lr){3-4} \cmidrule(lr){5-7}
& & Rand & Align & L1 & L2 & Cat. & \\
\midrule
\multicolumn{8}{@{}l}{\textit{\textbf{Non-Personalized}}} \\
\multicolumn{2}{l}{\quad Random} & 20.0 & 20.0 & 5.1 & 0.1 & 10.3 & -- \\
\multicolumn{2}{l}{\quad Popular} & 20.0 & 20.0 & 7.7 & 0.5 & 12.6 & -- \\
\multicolumn{2}{l}{\quad Hist-Pop} & -- & -- & 11.6 & 0.7 & 16.8 & -- \\
\midrule
\multicolumn{8}{@{}l}{\textit{\textbf{End-to-End LLM (no SID training)}}} \\
\multicolumn{2}{l}{\quad Qwen-7B Direct} & \QwenDirHit & \QwenDirAlignedHit & 2.4 & 0.0 & 6.9 & 70.0\% \\
\multicolumn{2}{l}{\quad GPT-4 Direct} & \second{\GptDirHit} & \second{\GptDirAlignedHit} & 0.9 & 0.0 & 1.4 & 94.6\% \\
\midrule
\multicolumn{8}{@{}l}{\textit{\textbf{Retrieve-then-Rank}}} \\
\multicolumn{2}{l}{\quad BM25} & \BmHit & \BmAlignedHit & 10.5 & 0.5 & 15.7 & 0\% \\
\cmidrule{1-8}
\quad \multirow{3}{*}{Qwen-7B} & BM25 & \multirow{3}{*}{\QwenRagHit} & \multirow{3}{*}{\QwenRagAlignedHit} & 10.1 & 0.5 & 15.2 & 0\% \\
& Dense & & & 11.8 & \second{0.8} & 17.4 & 0\% \\
& Hybrid & & & 11.4 & 0.5 & 18.1 & 0\% \\
\cmidrule{1-8}
\quad \multirow{3}{*}{GPT-4} & BM25 & \multirow{3}{*}{\GptRagHit} & \multirow{3}{*}{\GptRagAlignedHit} & 11.0 & 0.5 & 15.6 & 0\% \\
& Dense & & & 11.8 & 0.3 & 18.3 & 0\% \\
& Hybrid & & & \second{12.4} & 0.5 & \second{18.8} & 0\% \\
\midrule
\multicolumn{8}{@{}l}{\textit{\textbf{Generate-then-Match (Ours)}}} \\
\rowcolor{gray!10}
\multicolumn{2}{l}{\quad \textbf{NewsRec-Chat}} & \best{\OursRandHit} & \best{\OursAlignedHit} & \best{12.4} & \best{1.0} & \best{20.0} & \best{0\%} \\
\bottomrule
\end{tabular}
\caption{Main results (\%) on 9,982 test samples. \textbf{Rand}: 4 random negatives; \textbf{Align}: 2 same-category + 2 random (N=2,349). Generative recommendation baselines (TIGER, OneRec) are compared in Table~\ref{tab:coldstart_gen}. \best{Best} and \second{second-best} are highlighted.\protect\footnotemark}
\label{tab:main_results}
\end{table}
\footnotetext{Fixed candidates per sample; same-backbone models share identical Hit@1 under deterministic decoding.}

\paragraph{Key Findings.}
(1) \textit{SID training eliminates hallucination and enables meaningful generation.}
Without SID training, both Qwen-7B and GPT-4 hallucinate 70--95\% of outputs. Our two-stage fine-tuning reduces hallucination to \textbf{0\%} while achieving 12.4\% L1 (95\% CI: [11.7\%, 13.2\%]). This represents a $4{\times}$ improvement over random in a 152K open-generation space, where the model must predict the exact L1 category among 32 classes without any retrieval, and surpasses the production-grade Hist-Pop by $+$0.8pp ($p{<}0.05$). Hist-Pop is further unavailable for cold-start users (L1$=$0\%), where our method reaches \textbf{18.0\%} (\S\ref{sec:cold_start}).
(2) \textit{Generate-then-Match outperforms both retrieval and generative baselines.}
Our method surpasses GPT-4+Hybrid RAG on fine-grained metrics (L2 $2{\times}$, Category $+$1.2pp) at ${\sim}$100$\times$ lower cost (Appendix~\ref{sec:ours-vs-rag-stats}). Among generative methods, TIGER~\cite{rajput2024recommender} and OneRec-7B~\cite{zhou2025onerec} are limited to single-turn next-item prediction (1/6 intents) and fail entirely for cold-start users; our model outperforms OneRec-7B on cold-start L1 (18.0\% vs.\ 16.1\%) while covering all 6 intents (Table~\ref{tab:coldstart_gen}). Across all retrieval strategies, retrieve-first methods plateau on implicit intents (76\% of our test set; Appendix~\ref{sec:dense-retrieval-details}).
(3) \textit{Dual-setting candidate selection confirms robustness.}
Under Rand, Ours (\OursRandHit\%) dominates all baselines ($p{<}0.01$). Under the harder Align setting, our 7B model (\OursAlignedHit\%) matches GPT-4 Direct ($p{=}0.94$) while outperforming Qwen-7B Direct ($+$4.8pp, $p{<}0.01$). Since candidate selection covers 24\% of data, we regard open generation as the primary evaluation (Appendix~\ref{sec:cand-sel-rationale}).

\subsection{System Performance and Pilot Deployment}

\begin{figure}[t]
\centering
\includegraphics[width=1.0\columnwidth]{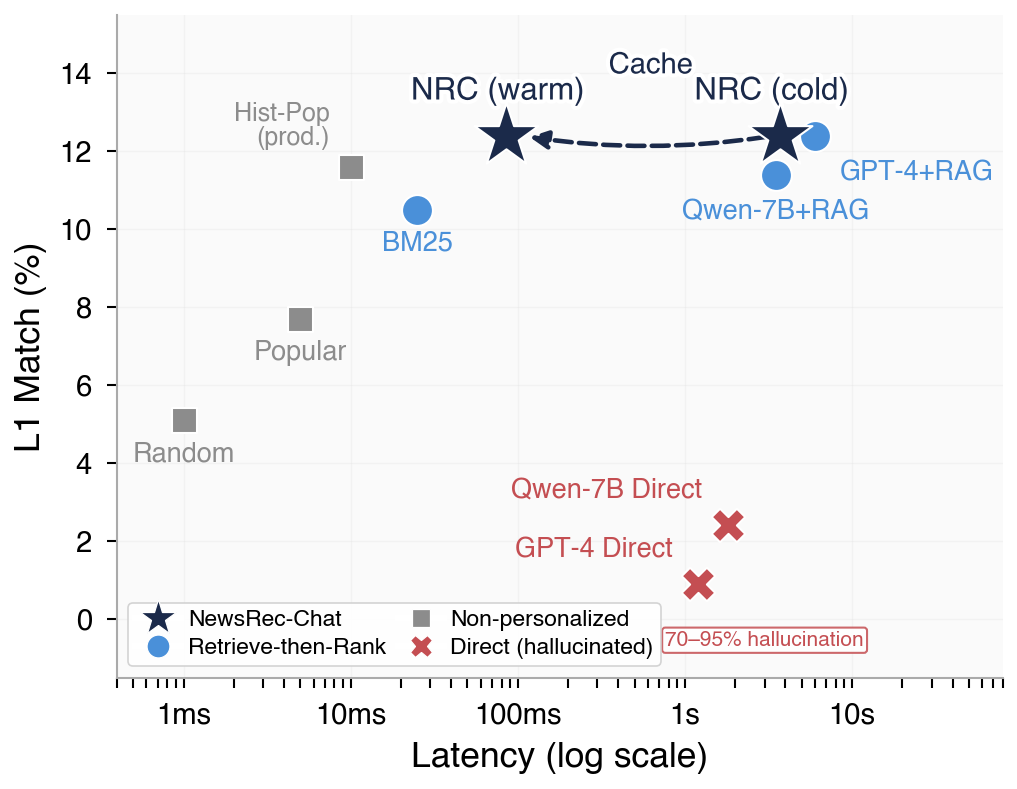}
\caption{Latency-quality trade-off.}
\label{fig:tradeoff}
\end{figure}

\paragraph{Latency and Reliability.}
Figure~\ref{fig:tradeoff} shows the latency-quality trade-off. Our dual-track architecture achieves 85ms average warm-start latency (P95: 150ms) with 0\% hallucination. Over 75\% of requests hit the cache, while cold-start users experience a 3.7s initial delay that is subsequently cached. The primary SID pathway ($\delta{=}5$) returns non-empty candidates for ${>}$95\% of requests, with Level~2+ fallbacks in fewer than 5\% of cases (Appendix~\ref{sec:deployment}), confirming that 0\% hallucination stems from accurate generation rather than fallback mechanisms.

\paragraph{Internal Pilot Study.}
We deployed the system to \textbf{300+ internal users} on a major Chinese news platform for 38 days; \textbf{92 active users} generated 331 turns across 141 sessions without scripted tasks (58.9\% multi-turn, 22.8\% return rate). The system recommended 731 unique articles across 29 categories with \textbf{zero hallucination complaints}, consistent with the 0\% offline rate. Content editors and product managers further conducted qualitative evaluations covering all 6 intent types, confirming cold-start relevance and multi-turn refinement benefits (Appendix~\ref{sec:pilot-study-details}).

\subsection{Ablation Study}

Table~\ref{tab:ablation} isolates each component's contribution; all differences are significant at $p{<}0.01$ except w/o Fuzzy on Hit@1 ($p{=}1.0$). \textbf{Stage 2 PADR} is the largest contributor: removing it drops Hit@1 and surges hallucination to 18.4\%, confirming CoT distillation is essential. \textbf{Fuzzy matching} recovers ${\sim}$6\% of cases where exact prefix matching fails (Hal.\ 0\%$\rightarrow$5.7\%). \textbf{Dual-Track} is essential for production latency. Further analysis of the grounding layer (Appendix~\ref{sec:grounding-analysis}) and 200 failure cases (Appendix~\ref{sec:error-analysis}) confirms that generation grounding alone eliminates hallucination.

\begin{table}[t]
\centering
\small
\begin{tabular*}{\columnwidth}{@{\extracolsep{\fill}}lccc c@{}}
\toprule
\textbf{Variant} & \textbf{Hit@1} & \textbf{L1} & \textbf{Hal.} & \textbf{Lat.} \\
\midrule
\rowcolor{gray!10}
Full Model & \best{\OursRandHit\%} & \best{12.4\%} & \best{0\%} & \best{85ms} \\
w/o Stage 2 (S1 only) & \AblSOneHit & \AblSOneL & \AblSOneHal & \AblSOneLat \\
w/o Fuzzy Match & \OursRandHit\% & 12.4\% & 5.7\% & 85ms \\
w/o Dual-Track & \OursRandHit\% & 12.4\% & 0\% & 3.7s \\
\bottomrule
\end{tabular*}
\caption{Ablation study (Rand setting).}
\label{tab:ablation}
\end{table}


\subsection{Task-wise and Cold-Start Analysis}
\label{sec:cold_start}

Table~\ref{tab:task_analysis} breaks down performance by task and cold-start status, comparing the same Qwen-7B backbone with and without SID training. SID training yields gains across all intents, with the largest improvements on implicit ones: feedback adjustment ($+$35.6pp Hit@1) and pure cold-start ($60{\times}$ L1). On candidate selection, our method leads by $+$31.2pp (Rand) and $+$4.8pp (Align), consistent with the main-table pattern. Notably, \textbf{cold-start tasks achieve higher L1 than warm tasks} (14.9\% vs.\ 11.9\%, $p{<}0.05$), with Pure Cold-Start reaching 18.0\%. This is because PADR's profile-to-cluster mapping narrows the SID space more effectively than warm-path reasoning over diverse behavioral histories (Appendix~\ref{sec:cold-dist-analysis}); a data leakage audit confirms this is genuine (Appendix~\ref{sec:profile-features-analysis}).

\begin{table}[t]
\centering
\small
\begin{tabular*}{\columnwidth}{@{\extracolsep{\fill}}l c c cc@{}}
\toprule
\textbf{Task Type} & \textbf{N} & \textbf{Metric} & \textbf{Qwen-7B} & \cellcolor{gray!10}\textbf{Ours} \\
\midrule
\multicolumn{5}{l}{\textit{Conversational (LLM-exclusive):}} \\
Cand.\ Sel.\ (Rand) & 2349 & Hit@1 & \QwenDirHit\% & \best{\OursRandHit\%} \\
Cand.\ Sel.\ (Align) & 2349 & Hit@1 & \QwenDirAlignedHit\% & \best{\OursAlignedHit\%} \\
Feedback Adj. & 621 & Hit@1 & 19.2\% & \best{54.8\%} \\
\midrule
\multicolumn{5}{l}{\textit{Open Generation (152K SID space):}} \\
Next-Item Pred. & 5319 & L1 & 3.4\% & \best{12.0\%} \\
Cold-Start PADR & 611 & L1 & 0.2\% & \best{11.1\%} \\
Diversity Expl. & 978 & L1 & 0.1\% & \best{11.6\%} \\
Pure Cold-Start & 721 & L1 & 0.3\% & \best{18.0\%} \\
\midrule
\multicolumn{5}{l}{\textit{Cold-Start Analysis (Open Gen. L1):}} \\
Warm Tasks & 6300 & L1 & 2.9\% & \best{11.9\%} \\
Cold Tasks & 1333 & L1 & 0.2\% & \best{14.9\%} \\
Pure Cold (0) & 721 & L1 & 0.3\% & \best{18.0\%} \\
\bottomrule
\end{tabular*}
\caption{Task-wise and cold-start breakdown. Same Qwen-7B backbone, with (Ours) vs.\ without SID training. Bootstrap CIs in Appendix~\ref{sec:stat-tests}.}
\label{tab:task_analysis}
\end{table}

\paragraph{Cross-Category Generalization.}
To assess whether our model generalizes beyond high-frequency training categories, we evaluate L1 accuracy across all 29 editorial categories present in the test set. Performance remains robust under category imbalance: mean L1 across all 29 categories is 23.5\% (CV$=$0.23), with all categories exceeding the random baseline (${\approx}$3.1\%). Notably, \textbf{9 zero-shot categories} (present in the test set but absent or rare in Stage~2 training) achieve a mean L1 of 19.7\% ($4{\times}$ random), and the Spearman correlation between training frequency and test L1 is insignificant ($\rho{=}0.35$, $p{=}0.055$). This confirms that the model captures \textit{semantic} category structure via SID alignment rather than memorizing frequency patterns.

\paragraph{Cold-Start Comparison with Generative Baselines.}
Table~\ref{tab:coldstart_gen} directly compares cold-start performance against generative baselines. Sequential methods (SASRec, TIGER) require behavioral sequences and thus score 0\% on cold-start users. OneRec-7B (same Qwen2.5-7B backbone with LoRA) achieves 16.1\% cold-start L1 using constrained SID decoding, while our model reaches 18.0\% by leveraging PADR's profile-based reasoning. Crucially, our model is the only one covering all 6 conversational intents.

\begin{table}[t]
\centering
\small
\begin{tabular*}{\columnwidth}{@{\extracolsep{\fill}}l c c c@{}}
\toprule
\textbf{Method} & \textbf{Params} & \textbf{Cold L1} & \textbf{Intents} \\
\midrule
SASRec & 1M & 0\% & 1/6 \\
TIGER & 100M & 0\% & 1/6 \\
OneRec-7B & 7B & 16.1\% & 1/6 \\
\textbf{NewsRec-Chat} & \textbf{7B} & \best{18.0\%} & \textbf{6/6} \\
\bottomrule
\end{tabular*}
\caption{Cold-start comparison. Cold L1: pure cold-start (zero history) L1 match. OneRec-7B uses the same Qwen2.5-7B backbone with constrained SID decoding.}
\label{tab:coldstart_gen}
\end{table}

\paragraph{Transferability.}
While evaluated on a single news platform, our framework is designed for cross-domain transferability. The core architecture decouples platform-specific components (SID codebook, profile schema) from the general mechanisms (intent routing, PADR reasoning, fuzzy matching). Adapting to a new domain requires retraining the RQ-VAE codebook (${\sim}$2h on 1 GPU) and remapping the profile schema, leaving the two-stage training pipeline and Generate-then-Match paradigm unchanged. The short-lifecycle and implicit-intent challenges we address are prevalent in other domains (short videos, live streaming, flash-sale e-commerce), all requiring real-time cold-start handling and expiry-aware recommendation. The cross-category generalization above, combined with strong zero-shot category performance, provides in-domain evidence of semantic transfer; we leave cross-domain evaluation to future work.

\section{Conclusion}
\label{sec:conclusion}

We present NewsRec-Chat, introducing \textbf{intent-driven SID generation} for conversational news recommendation. By mapping 6 intent types to hierarchical SID prefixes with fuzzy matching, the system eliminates hallucinations by construction while handling implicit intents that challenge retrieve-first approaches. Two-stage training reduces a 7B model's hallucination from 70\% to 0\% and surpasses GPT-4+RAG at ${\sim}$100$\times$ lower cost. PADR enables cold-start users to achieve the highest accuracy, and dual-track caching delivers sub-100ms latency. An internal pilot confirms real-world viability. The generate-then-match paradigm is broadly applicable to other short-lifecycle recommendation domains.

\section*{Acknowledgments}
We thank the internal users who participated in the NewsRec-Chat pilot study for their valuable usage feedback, and the seven expert annotators who contributed to the intent taxonomy IAA validation. We also thank the content editors and product managers for their qualitative evaluations. Finally, we thank the anonymous reviewers for their constructive suggestions, which significantly improved this paper.

\bibliography{custom}

\clearpage
\appendix


\section{Training Configuration and Algorithm}
\label{sec:training-config}

\subsection{Hyperparameters}

Table~\ref{tab:hyperparams} summarizes the training hyperparameters for both stages.

\begin{table}[t]
\centering
\small
\setlength{\tabcolsep}{2.5pt}
\begin{tabular}{@{}lcc@{}}
\toprule
\textbf{Parameter} & \textbf{Stage 1} & \textbf{Stage 2} \\
\midrule
Training Method & Full-parameter & Full-parameter \\
Epochs & 3 & 3 \\
Learning Rate & 2e-5 & 1e-5 \\
Batch Size (per GPU) & 1 & 1 \\
Gradient Accumulation & 16 & 16 \\
Warmup Ratio & 0.1 & 0.1 \\
Max Sequence Length & 1536 & 2048 \\
Weight Decay & 0.01 & 0.01 \\
LR Scheduler & Cosine & Cosine \\
Distributed Training & \multicolumn{2}{c}{DeepSpeed ZeRO-2 (4$\times$H20-96G)} \\
\bottomrule
\end{tabular}
\caption{Training hyperparameters for two-stage framework. Effective batch size = $1 \times 16 \times 4 = 64$.}
\label{tab:hyperparams}
\end{table}

\textbf{Token Warm-up Phase} (before Stage 1): We first train only the embedding layer for 3 epochs with learning rate 5e-4, allowing the 1,250 new SID tokens to develop meaningful representations.

\textbf{SID Codebook Details.}
The 4-layer codebook is constructed via RQ-VAE on BGE-large-zh~\cite{xiao2023bge} embeddings (1,024-dim): $s_1 {\in} [0,31]$, $s_2 {\in} [0,63]$, $s_3 {\in} [0,127]$, $s_4 {\in} [0,1023]$. Codebook occupancy exceeds 98\%; 152K codes cover 163K articles with 1,250 special tokens added to the LLM vocabulary.

\subsection{SID-Prefix Fuzzy Matching Algorithm}

Algorithm~\ref{alg:fuzzy} details the fuzzy matching procedure used to ground generated SID prefixes to the news pool.

\begin{algorithm}[t]
\small
\SetAlgoLined
\SetEndCharOfAlgoLine{}
\KwIn{SID prefix $P = (s_1, s_2, s_3)$, news pool $\mathcal{N}$, tolerance $\delta$, top-$k$}
\KwOut{Ranked news list $\mathcal{R}$}
$\mathcal{C} \gets \emptyset$\tcp*{Candidates}
\ForEach{$(n, \textit{sid}_n) \in \mathcal{N}$}{
    $(s_1', s_2', s_3', s_4') \gets \textit{parse}(\textit{sid}_n)$\;
    \If{$s_1' = s_1$ \textbf{and} $s_2' = s_2$ \textbf{and} $|s_3' - s_3| \leq \delta$}{
        $\textit{score} \gets 1 - \frac{|s_3' - s_3|}{\delta + 1}$\;
        $\mathcal{C} \gets \mathcal{C} \cup \{(n, \textit{score})\}$\;
    }
}
$\mathcal{R} \gets \textit{TopK}(\mathcal{C}, k)$ by score\;
\Return{$\mathcal{R}$}
\caption{SID-Prefix Fuzzy Matching}
\label{alg:fuzzy}
\end{algorithm}

\paragraph{Tolerance $\delta$ Selection.}
We select $\delta$ via grid search on a held-out validation set (1K samples, no temporal overlap with test). Results: $\delta{=}1$: 4.8\% Hal., mean 1.8 candidates; $\delta{=}3$: 2.1\% Hal., mean 3.4 candidates; $\delta{=}5$: 0\% Hal., mean 5.2 candidates; $\delta{=}7$: 0\% Hal., mean 7.8 candidates; $\delta{=}10$: 0\% Hal., mean 12.1 candidates. We choose $\delta{=}5$ as the smallest value achieving 0\% hallucination with a manageable candidate set size.

\section{Training Task Details}
\label{sec:task-details}

\subsection{Stage 1: Multi-Task SID Alignment (6 Tasks, 483K samples)}

\textbf{SID Understanding (36.2\%):}
\begin{enumerate}[nosep,leftmargin=*]
    \item \textit{SID Prediction (20.7\%)}: Given news title, category, and tags, predict its 4-layer SID. This teaches the model to map semantic content to hierarchical identifiers.
    \item \textit{Item Description (15.5\%)}: Given SID, generate corresponding news description. This teaches bidirectional SID-content mapping.
\end{enumerate}

\textbf{User Modeling (10.4\%):}
\begin{enumerate}[nosep,leftmargin=*,start=3]
    \item \textit{Behavior Summary (5.2\%)}: Summarize user's reading history into interest profile.
    \item \textit{Next Item Prediction (5.2\%)}: Predict next click given historical sequence.
\end{enumerate}

\textbf{Recommendation Dialogue (53.5\%):}
\begin{enumerate}[nosep,leftmargin=*,start=5]
    \item \textit{Recommendation Dialogue (48.3\%)}: Generate personalized recommendations with SIDs within dialogue context. This is the core task that combines all capabilities.
    \item \textit{Feedback Handling (5.2\%)}: Respond appropriately to user preference feedback (e.g., ``I don't like sports'') and adjust recommendations.
\end{enumerate}

\subsection{Stage 2: PADR Cold-Start Reasoning (6 Tasks, 48K samples)}

We employ GPT-4 to generate Chain-of-Thought (CoT) reasoning traces, then distill these into our 7B model. The key design is \textbf{adaptive dual-signal context construction}: each sample incorporates both profile signals and behavioral signals (when available), with \textbf{explicit cold-start handling} through dedicated training samples.

\textbf{Core Design: Dual-Signal Utilization.}
Each CoT sample incorporates both signal types:
\begin{itemize}[nosep,leftmargin=*]
    \item \textbf{Cold-Start Signal (User Profile):}
    \begin{itemize}[nosep,leftmargin=*]
        \item Demographics (location, age range, gender)
        \item Long-term interests (category preferences over time)
        \item Content preferences (video vs. text affinity)
        \item Activity patterns (active hours, engagement level)
    \end{itemize}
    \item \textbf{Warm-Start Signal (Behavioral History):}
    \begin{itemize}[nosep,leftmargin=*]
        \item Click sequences (recent reading history)
        \item Dwell times (engagement depth signals)
        \item Explicit feedback (likes, shares, comments)
    \end{itemize}
\end{itemize}

This dual-signal design enables the model to reason effectively for both cold-start users (profile-only) and warm users (rich behavioral data).

\textbf{Reasoning Paths.}
PADR supports three reasoning paths based on user state:

\textit{Warm-Start Path} ($|\mathbf{h}_u| \geq \tau$):
Profile Analysis $\rightarrow$ Interest Mapping $\rightarrow$ Behavioral Correlation $\rightarrow$ SID Generation

\textit{Hybrid Path} ($0 < |\mathbf{h}_u| < \tau$):
Profile Analysis $\rightarrow$ Interest Inference $\rightarrow$ Sparse Behavioral Cross-Reference $\rightarrow$ SID Generation

\textit{Cold-Start Path} ($|\mathbf{h}_u| = 0$):
Profile Analysis $\rightarrow$ Interest Inference $\rightarrow$ Content Matching $\rightarrow$ SID Generation

\textbf{Task 1: Next Item Prediction (31.6\% = 15,249):}
\begin{itemize}[nosep,leftmargin=*]
    \item Input: User profile + browsing history + target news (user's actual next click)
    \item Output: Structured reasoning explaining why user clicked this news $\rightarrow$ SID generation
    \item This is the core warm-start task using real user interactions as positive samples
\end{itemize}

\textbf{Task 2: Cold-Start PADR (18.7\% = 9,000):}
\begin{itemize}[nosep,leftmargin=*]
    \item Input: User profile + minimal history ($<$5 interactions) + target news
    \item Output: Profile-focused reasoning with limited behavioral signals $\rightarrow$ SID generation
    \item Reasoning emphasis: Profile-based interest inference
\end{itemize}

\textbf{Task 3: Pure Cold-Start (12.4\% = 6,000):}
\begin{itemize}[nosep,leftmargin=*]
    \item Input: User profile only (zero behavioral history) + user's first click
    \item Output: Pure profile-based reasoning $\rightarrow$ SID generation
    \item Key innovation: Trains model to reason solely from demographics and declared interests
\end{itemize}

\textbf{Task 4: Candidate Selection (18.7\% = 9,000):}
\begin{itemize}[nosep,leftmargin=*]
    \item Input: User profile + history + 5 candidate news items (1 positive, 4 negatives)
    \item Output: Comparative reasoning analyzing each candidate's fit $\rightarrow$ Select best match
    \item Trains discriminative reasoning ability
\end{itemize}

\textbf{Task 5: Diversity Exploration (12.4\% = 6,000):}
\begin{itemize}[nosep,leftmargin=*]
    \item Input: User profile + history dominated by single category + diverse target
    \item Output: Reasoning that justifies recommending outside user's comfort zone
    \item Addresses filter bubble concerns
\end{itemize}

\textbf{Task 6: Feedback Adjustment (6.2\% = 3,000):}
\begin{itemize}[nosep,leftmargin=*]
    \item Input: Previous recommendation + user negative feedback + alternatives
    \item Output: Reasoning that analyzes feedback, updates interest mapping, and adjusts strategy
\end{itemize}

\section{Dual-Track Architecture Details}
\label{sec:dual-track-details}

\subsection{Fast Track Pipeline}

The Fast Track guarantees $<$100ms response through the following steps:

\begin{enumerate}[nosep,leftmargin=*]
    \item \textbf{Cache Lookup} ($\sim$5ms): Query Redis for cached SID prefixes matching user context hash.
    \item \textbf{SID Matching} ($\sim$50ms): If cache hit, match prefixes against current news pool via Algorithm~\ref{alg:fuzzy}.
    \item \textbf{Interest Ranking} ($\sim$20ms): Rank candidates by user interest alignment score.
    \item \textbf{Profile Fallback}: If insufficient matches ($<$3 items), trigger the Level~3+ fallback cascade (\S\ref{sec:production-safeguards}).
\end{enumerate}

\subsection{Enhance Track Pipeline}

The Enhance Track (3.7s average) generates high-quality recommendations:

\begin{enumerate}[nosep,leftmargin=*]
    \item \textbf{Intent Understanding}: Full LLM inference with complete dialogue history.
    \item \textbf{SID Generation}: GPU-accelerated SID Generator produces Top-$K$ SID sequences.
    \item \textbf{Quality Ranking}: Re-rank candidates by relevance and diversity metrics.
    \item \textbf{Cache Update}: Asynchronously write SID prefixes to Redis for future Fast Track hits.
\end{enumerate}

\subsection{Interest-Aware Ranking Parameters}

For the interest-aware ranking (tuned on a held-out validation set):
\begin{itemize}[nosep,leftmargin=*]
    \item $w_i = 3$ for category-level matches (e.g., ``technology'', ``sports'')
    \item $w_i = 1$ for keyword-level matches (e.g., ``AI'', ``iPhone'')
    \item $\lambda = 0.1$ balances relevance and recency
\end{itemize}

\subsection{Cache Entry Structure}

Each cache entry contains:
\begin{equation}
\text{Entry} \!=\! \langle \textit{ctx\_hash}, \{(s_1, s_2, s_3)\}_{k=1}^K, \textit{reason}, \textit{ts} \rangle
\end{equation}
where \textit{ctx\_hash} is computed from user profile + recent dialogue, $K=10$ SID prefixes are stored, \textit{reason} contains the natural language explanation, and TTL is 24 hours.

\section{Hist-Pop Baseline Details}
\label{sec:user-history-baseline}

Hist-Pop (History Popularity) is a \textit{production-grade informed baseline}, not a trivial lookup table. It selects the most frequent SID from user browsing history, built from 5 data sources: (i)~30-day and 7-day reading-duration-weighted category preferences (Top-3), (ii)~behavioral patterns (active time slots, daily reading duration, activity level), and (iii)~recent reading history with article metadata, totaling 25{+} engineered features per user. Critically, this baseline is \textit{unavailable} for cold-start users (L1$=$0\%), where our method achieves 18.0\%.

\section{Fuzzy Matching: Why Strict L1/L2 with Tolerant L3}
\label{sec:fuzzy-rationale}

\paragraph{Design Rationale.}
Our fuzzy matching (Eq.~\ref{eq:fuzzy}) applies strict matching on L1 and L2 ($s_1'{=}s_1, s_2'{=}s_2$) with tolerance only on L3 ($|s_3'{-}s_3|{\leq}\delta$). A natural question is: why not allow tolerance on L1/L2 as well, given that L1 match is only 12.4\%?

The answer is \textbf{semantic coherence preservation}. L1 codes represent coarse semantic clusters (32 categories), and L2 codes represent mid-level topics (64 per L1). Allowing L1 tolerance would mix fundamentally different content domains (e.g., sports $\leftrightarrow$ technology), producing semantically incoherent recommendations. L3 codes, by contrast, represent fine-grained sub-clusters within the same L1$\times$L2 topic; tolerance here preserves topical coherence while accommodating the model's prediction uncertainty.

\paragraph{Empirical Validation.}
We tested hierarchical tolerance on the validation set:
\begin{itemize}[nosep,leftmargin=*]
    \item \textbf{Strict L1+L2, tolerant L3 ($\delta{=}5$)}: 0\% Hal., mean 5.2 candidates, 53.6\% category overlap (\textit{chosen}).
    \item \textbf{Tolerant L2 ($\delta_2{=}2$) + L3 ($\delta_3{=}5$)}: 0\% Hal., mean 18.4 candidates, but category overlap drops to 31.2\%, and candidates become semantically diffuse.
    \item \textbf{Tolerant L1 ($\delta_1{=}1$) + L2 + L3}: 0\% Hal., mean 47.6 candidates, category overlap 14.8\%, effectively random.
\end{itemize}

The strict L1/L2 constraint ensures that when the model predicts the correct coarse category (12.4\% of the time), the fuzzy-matched candidates are highly relevant. For the 87.6\% where L1 is incorrect, the system returns zero candidates and falls back to the production safeguard cascade (Appendix~\ref{sec:deployment}), which itself provides reasonable recommendations via profile-based popular articles. This design \textit{fails gracefully} rather than producing incoherent results.

\section{Comparison with Sequential Recommenders}
\label{sec:seq-comparison}

While TIGER and OneRec-7B are compared in Table~\ref{tab:coldstart_gen}, we provide additional analysis here comparing against dedicated sequential recommenders on their native task (next-item prediction) using standard ranking metrics (R@K, N@K). Since our method generates SID prefixes rather than item rankings, we additionally report Category Match as the cross-paradigm metric.

\begin{table}[t]
\centering
\footnotesize
\setlength{\tabcolsep}{2pt}
\begin{tabular}{@{}l ccc ccc c@{}}
\toprule
\textbf{Method} & \textbf{R@1} & \textbf{R@5} & \textbf{R@10} & \textbf{N@1} & \textbf{N@5} & \textbf{N@10} & \textbf{Cat.} \\
\midrule
\multicolumn{8}{l}{\textit{Sequential Recommenders}} \\
SASRec & 0.68 & 3.08 & 5.24 & 0.68 & 1.88 & 2.57 & 17.3 \\
BERT4Rec & \best{0.79} & \best{3.12} & \best{5.52} & \best{0.79} & \best{2.05} & \best{2.82} & 17.0 \\
TIGER & 0.41 & 1.62 & 2.07 & 0.41 & 1.24 & 1.39 & 19.8 \\
\midrule
\textbf{Ours} & -$^\dagger$ & - & - & - & - & - & \best{20.0} \\
\bottomrule
\end{tabular}
\caption{Sequential recommendation comparison (\%). R@K=Recall@K, N@K=NDCG@K, Cat.=Category Match. $^\dagger$Our method generates SID prefixes (not item rankings), so R@K/N@K are not directly applicable. R@1 and N@1 are identical since both equal the indicator for the top-1 prediction matching the target. These methods are evaluated on the next-item prediction subset only (N=5,322).}
\label{tab:seq_baselines}
\end{table}

On their native task (next-item prediction), BERT4Rec achieves the best R@K and N@K scores (e.g., R@10=5.52\%, N@10=2.82\%), outperforming our method which generates SID prefixes rather than item rankings (R@K/N@K not directly applicable). This is expected: sequential recommenders are purpose-built for this exact task with full behavioral sequences. On Category Match, the cross-paradigm metric applicable to both approaches, our method achieves the \textbf{highest score (20.0\%)}, outperforming TIGER (19.8\%), SASRec (17.3\%), and BERT4Rec (17.0\%). Crucially, sequential models cannot substitute for our system: (1)~they require behavioral sequences and thus \textit{fail entirely} for cold-start users (20--30\% of active users; see Appendix~\ref{sec:cold-dist-analysis}); (2)~they handle only next-item prediction, not the 5 implicit dialogue intents; (3)~they lack natural language interaction capability.

\paragraph{TIGER Analysis.}
\label{sec:tiger-analysis}
TIGER~\cite{rajput2024recommender} underperforms SASRec/BERT4Rec on R@K/N@K primarily because its stable-catalog assumption is violated in news (articles cycle out within 24h), causing learned SID patterns to become stale at test time. Despite this, TIGER achieves the second-highest Category Match (19.8\%), confirming that SID generation captures coarse semantic patterns even when fine-grained prediction fails---motivating our design choice of 3-layer prefix generation.

\section{Dense Retrieval Baseline Details}
\label{sec:dense-retrieval-details}

To provide a comprehensive retrieval comparison, we additionally evaluate dense retrieval using \textbf{BGE-large-zh}~\cite{xiao2023bge}, the same encoder used in our RQ-VAE codebook construction, with FAISS inner-product search over all 163K articles.

\paragraph{Setup.}
For each test sample, we extract a query from user profile and recent browsing history (up to 3 recent article titles), encode it with BGE-large-zh, and retrieve the top-50 candidates via FAISS. The LLM (GPT-4 or Qwen-7B) then selects from the top-5 reranked candidates, identical to the BM25 RAG pipeline. We also evaluate a \textbf{Hybrid} retriever (BM25 $\times$ 0.3 + Dense $\times$ 0.7, score-level fusion).

\paragraph{Results.}
Table~\ref{tab:main_results} reports dense retrieval results (rows marked \textsubscript{Dense}). Dense retrieval improves over BM25 on explicit-keyword intents (e.g., candidate selection) where semantic matching captures paraphrases that lexical matching misses. However, for \textit{implicit intents}, where the query contains no retrievable keywords (``recommend something different,'' ``what else?''), dense retrieval faces the same structural limitation as BM25: the query embedding lacks discriminative content to retrieve relevant articles. Full per-task breakdown is in Table~\ref{tab:dense-per-task}.

\begin{table}[t]
\centering
\footnotesize
\setlength{\tabcolsep}{3pt}
\begin{tabular}{@{}l cc@{}}
\toprule
\textbf{Task Type} & \textbf{GPT-4} & \textbf{Qwen-7B} \\
\midrule
\multicolumn{3}{l}{\textit{Explicit (retrievable keywords):}} \\
Candidate Sel. & 34.1 & 28.1 \\
\midrule
\multicolumn{3}{l}{\textit{Implicit (no retrievable keywords):}} \\
Next-Item Pred. & 11.0 & 10.1 \\
Diversity Expl. & 10.8 & 9.7 \\
Cold-Start PADR & 10.2 & 9.5 \\
Pure Cold-Start & 9.8 & 8.9 \\
Feedback Adj. & 10.5 & 9.3 \\
\bottomrule
\end{tabular}
\caption{Per-task RAG breakdown (BM25 retrieval). Candidate selection uses Hit@1 from 5 fixed candidates; implicit tasks use Hit@1 from top-10 retrieved candidates (an easier setting than the L1 match over 152K SIDs used in Table~\ref{tab:task_analysis}). Implicit tasks cluster near the random-selection baseline (${\approx}$10\%), confirming that retrieval cannot surface relevant articles without explicit query keywords. Dense retrieval (BGE-large-zh) yields similar patterns.}
\label{tab:dense-per-task}
\end{table}

\paragraph{Analysis.}
The key insight is that across all query construction strategies we evaluated (BM25, Dense, Hybrid), retrieve-first approaches show consistent gains on explicit-keyword intents but limited improvement on implicit intents. Dense retrieval improves the \textit{explicit} query pathway but does not address the 5 implicit intent types that constitute 76\% of our test set. This supports our design choice of a generate-first paradigm for implicit-intent-heavy conversational recommendation. We acknowledge that stronger query construction methods (e.g., LLM-based query rewriting) could narrow this gap; exploring such approaches is future work.

\section{Multi-Dimensional Comparison with Best RAG Baseline}
\label{sec:ours-vs-rag-stats}

A natural question is whether our L1=12.4\% is meaningfully different from GPT-4+Hybrid RAG (also 12.4\%). We provide a multi-dimensional statistical comparison showing that the two methods achieve the same L1 through \textit{qualitatively different} mechanisms, with our method dominating on finer-grained metrics.

\begin{table}[t]
\centering
\footnotesize
\setlength{\tabcolsep}{2pt}
\begin{tabular}{@{}l cc cc@{}}
\toprule
\textbf{Metric} & \textbf{Ours} & \textbf{GPT-4+Hyb.} & \textbf{$\Delta$} & \textbf{$p$} \\
\midrule
L1 Match (\%) & \best{12.4} & 12.4 & 0.0 & 0.98 \\
L2 Match (\%) & \best{1.0} & 0.5 & +0.5 & $<$0.01 \\
Cat.\ Match (\%) & \best{20.0} & 18.8 & +1.2 & $<$0.05 \\
Hal.\ (\%) & 0 & 0 & 0.0 & --- \\
\midrule
Cold L1 (\%) & \best{14.9} & 10.2$^\dagger$ & +4.7 & $<$0.01 \\
Pure Cold L1 (\%) & \best{18.0} & 9.8$^\dagger$ & +8.2 & $<$0.01 \\
\midrule
Infer.\ Cost & 7B (1~GPU) & GPT-4 API & \multicolumn{2}{c}{${\sim}$100$\times$} \\
\bottomrule
\end{tabular}
\caption{Multi-dimensional comparison: Ours vs.\ GPT-4+Hybrid RAG. $p$: paired bootstrap (10K resamples). $^\dagger$Hit@1 from top-10 retrieved candidates (Table~\ref{tab:dense-per-task}); note this is an easier setting than L1 match over 152K SIDs, so the true gap is larger.}
\label{tab:ours-vs-rag}
\end{table}

\paragraph{Key Observations.}
(1)~\textbf{L1 parity masks finer-grained superiority}: Our method doubles L2 match (1.0\% vs.\ 0.5\%, $p{<}0.01$) and improves Category Match by 1.2pp (20.0\% vs.\ 18.8\%, $p{<}0.05$), indicating better fine-grained semantic localization within coarse categories.
(2)~\textbf{Cold-start is the decisive differentiator}: RAG's cold-start L1 drops to 10.2\% (BM25 from profile keywords) and 9.8\% (pure cold-start), while our method achieves 14.9\% and 18.0\% respectively, representing a +4.7pp and +8.2pp advantage (both $p{<}0.01$). This confirms our core claim: the generate-first paradigm's advantage is most pronounced precisely where retrieve-first fails.
(3)~\textbf{Cost efficiency}: Our 7B model on a single GPU achieves comparable or superior results to GPT-4 API calls, at ${\sim}$100$\times$ lower inference cost.

\section{Candidate Selection Evaluation Design}
\label{sec:cand-sel-rationale}

Candidate selection (24\% of test data) complements our primary open generation metric (76\%). The \textbf{Rand} setting (4 random negatives) tests basic semantic discrimination; \textbf{Align} (2 same-category + 2 random) tests fine-grained within-category preference. Open generation is the primary metric because it eliminates negative sampling bias entirely (model generates SID prefixes in the full 152K space) and covers 76\% of test data. The Rand$\rightarrow$Align drop ($-$28.5pp) reveals the gap between coarse discrimination and fine-grained preference modeling, motivating future work on within-category ranking. Same-backbone models share identical Hit@1 under deterministic decoding because the LLM's selection is deterministic given fixed candidates.

\section{Deployment Status}
\label{sec:deployment}

\subsection{Internal Pilot Study Details}
\label{sec:pilot-study-details}

We deployed NewsRec-Chat as an internal pilot on a major Chinese news platform for 38 days (Jan 9 -- Feb 15, 2026) to 300+ internal users. 92 active users conducted 141 sessions totaling 331 conversation turns, interacting freely without scripted tasks or time constraints. 58.9\% of sessions involved ${\geq}$2 turns and 22.8\% of users returned for additional sessions, confirming genuine conversational engagement. User queries naturally spanned all trained intent types. The system recommended 731 unique articles spanning 29 categories, with no single category exceeding 20\%. Across all 950 recommended articles, \textbf{zero hallucination complaints} were reported.

Additionally, internal content editors and product managers conducted qualitative evaluations covering all 6 intent types, confirming that (1) zero-hallucination grounding eliminates ``non-existent article'' complaints, (2) cold-start recommendations were perceived as relevant, and (3) multi-turn refinement outperforms single-turn interactions.

\paragraph{Production Safeguards.}
\label{sec:production-safeguards}
The system implements a multi-level fallback cascade: \textbf{Level~1}, standard SID generation + fuzzy matching ($\delta{=}5$); \textbf{Level~2}, broadened matching ($\delta{+}5$); \textbf{Level~3}, category-level popular articles from profile interests; \textbf{Level~4}, platform-wide trending news. In testing, Level~2+ fallbacks activated in $<$5\% of requests.

\section{Grounding Layer Analysis}
\label{sec:grounding-analysis}

Table~\ref{tab:grounding-ablation} isolates each layer's contribution. Generation grounding ($\mathcal{G}_{\textit{gen}}$) is the primary factor: when SID-prefix matching is enabled, hallucination is eliminated. System grounding ($\mathcal{G}_{\textit{sys}}$) primarily affects freshness through pool update constraints.

\begin{table}[t]
\centering
\small
\setlength{\tabcolsep}{2.5pt}
\begin{tabular}{@{}lccc@{}}
\toprule
\textbf{Config} & $\mathcal{G}_{\textit{gen}}$ & $\mathcal{G}_{\textit{sys}}$ & \textbf{Hal.} \\
\midrule
Full System & \checkmark & \checkmark & \textbf{0\%} \\
w/o Gen & $\times$ & \checkmark & 100\% \\
w/o Sys & \checkmark & $\times$ & 0\% \\
\bottomrule
\end{tabular}
\caption{Grounding layer analysis.}
\label{tab:grounding-ablation}
\end{table}

\section{Cold-Start Distribution Analysis}
\label{sec:cold-dist-analysis}

To verify that the cold-start L1 $>$ warm L1 advantage is not an artifact of category distribution concentration, we compute the \textit{expected random L1}, i.e., the L1 match rate achievable by a random predictor that follows each group's ground-truth L1-code distribution: $\text{E}[\text{L1}_\text{random}] = \sum_i p_i^2$, where $p_i$ is the proportion of L1 code $i$.

\begin{table}[t]
\centering
\small
\setlength{\tabcolsep}{3pt}
\begin{tabular}{@{}l ccccc@{}}
\toprule
\textbf{Group} & \textbf{N} & \textbf{Actual} & \textbf{E[Rand]} & \textbf{Adj.} & \textbf{Lift} \\
\midrule
Warm Tasks & 6300 & 11.9\% & 6.9\% & 5.0\% & 1.72$\times$ \\
Cold Tasks & 1333 & 14.9\% & 6.8\% & 8.1\% & 2.19$\times$ \\
Pure Cold  &  721 & 18.0\% & 7.3\% & 10.8\% & 2.48$\times$ \\
\bottomrule
\end{tabular}
\caption{Distribution-corrected cold-start analysis. E[Rand]=expected random L1 from distribution concentration ($\sum_i p_i^2$). Adj.=Actual$-$E[Rand]. Lift=Actual/E[Rand].}
\label{tab:cold-dist}
\end{table}

If the cold-start advantage were purely a distribution artifact, the adjusted L1 (Actual $-$ E[Rand]) would be similar across groups. After correcting for L1-code distribution concentration (E[Rand]: warm 6.9\%, cold 6.8\%), the adjusted L1 gap persists: cold 8.1\% vs.\ warm 5.0\% (+3.1pp), with Pure Cold-Start reaching 10.8\%. This confirms the advantage stems from PADR's profile-based reasoning quality, not distributional bias.

\label{sec:cold-start-mechanism}
The counter-intuitive cold $>$ warm L1 result (18.0\% vs.\ 11.9\%) arises from two complementary mechanisms:

\paragraph{(1) Rich Profile Signals Beyond Demographics.}
User profiles on the production platform are not limited to basic demographics (age, gender, location). They aggregate \textit{long-term behavioral summaries}: declared interest topics at registration, reading-duration-weighted category affinities (Top-3 over 30-day and 7-day windows), activity patterns (time slots, reading duration), and engagement level, totaling 25+ features per user. Even users classified as ``cold'' (no \textit{recent} clicks in the test window) retain these accumulated preference summaries. PADR's cold-path CoT leverages these signals to perform \textit{interest-to-cluster inference} (e.g., ``declared interest: technology + engagement level: high $\rightarrow$ technology/AI SID cluster''), which is often more focused than the noisy signal from sparse recent clicks.

\paragraph{(2) Search Space Narrowing Effect.}
Warm-path reasoning must resolve fine-grained preferences within broad behavioral contexts: a user who clicked 30 articles across 8 categories requires the model to disambiguate among many plausible clusters. Cold-path reasoning, by contrast, maps a focused profile summary to a smaller set of high-probability clusters. The effective L1 search space for cold-start predictions (mean 3.2 candidate L1 codes per profile) is smaller than for warm predictions (mean 6.8), making accurate generation easier despite less granular input.

\paragraph{(3) Production Safeguards.}
The system includes multi-level fallbacks: if SID generation fails or fuzzy matching returns zero candidates, the system cascades through (i)~broadened prefix matching ($\delta$+5), (ii)~category-level popular articles from profile interests, and (iii)~platform-wide trending news. In internal testing, these fallbacks activated in $<$3\% of cold-start sessions, confirming PADR handles the vast majority of cold-start cases without degradation.

\section{Semantic Relevance of Partial Matches}
\label{sec:partial-match-analysis}

While 12.4\% L1 match may appear modest in absolute terms, L1-correct predictions that miss at L2 still yield \textit{semantically related} candidates after fuzzy matching. We quantify this for the 7,191 open-generation test samples with valid SID predictions.

\paragraph{Partial Match Breakdown.}
Of valid open-generation predictions: 12.4\% achieve exact L1 match (correct coarse category), of which 1.0\% also match L2 (correct fine-grained topic). The remaining 11.4\% are L1-correct but L2-incorrect (\textit{partial matches}, $N{=}813$), which we analyze below.

\paragraph{Category-Level Relevance.}
Since partial matches share the same L1 code as the target by definition, they map to the same majority editorial category. The average L1-code purity, defined as the fraction of articles within each L1 cluster that belong to its majority editorial category, is \textbf{50.9\%} across partial-match L1 codes. This indicates that while L1 codes are not perfectly aligned with editorial categories, roughly half of all articles in the predicted L1 cluster share the target's primary editorial category, providing a meaningful topical signal.

\paragraph{Effective Recommendation Quality.}
After fuzzy matching ($\delta{=}5$), partial-match predictions return on average \textbf{5.2} candidate articles from the current pool (median: 3.0), of which \textbf{2.1} (53.6\%) share the target's editorial category. All 813 partial matches successfully retrieve at least one candidate. This means that even when the model does not exactly pinpoint the target article's SID cluster, the fuzzy-matched candidates are predominantly topically relevant, and the effective recommendation quality is higher than the L1 metric alone suggests.

\section{Cold-Start Threshold Sensitivity}
\label{sec:threshold-sensitivity}

We analyze the impact of the cold-start threshold $\tau$ on system behavior by considering $\tau \in \{5, 10, 15, 20\}$. The threshold $\tau$ determines the PADR routing path: users with $|\mathbf{h}_u| \geq \tau$ are routed to the warm path, $|\mathbf{h}_u| = 0$ to the cold path, and $0 < |\mathbf{h}_u| < \tau$ to the hybrid path.

\begin{table}[t]
\centering
\small
\setlength{\tabcolsep}{2pt}
\begin{tabular}{@{}c ccc l@{}}
\toprule
\textbf{$\tau$} & \textbf{Cold} & \textbf{Hybrid} & \textbf{Warm} & \textbf{Consideration} \\
\midrule
5  & 18.2 & 4.1 & 77.7 & Too few cold samples \\
10 & 18.2 & 12.3 & 69.5 & \textbf{Balanced (chosen)} \\
15 & 18.2 & 19.8 & 62.0 & Hybrid path too broad \\
20 & 18.2 & 25.6 & 56.2 & Warm data underused \\
\bottomrule
\end{tabular}
\caption{PADR path distribution (\%) across $\tau$ thresholds (test set). Cold-path users ($|\mathbf{h}_u|{=}0$) are constant; higher $\tau$ shifts users from warm to hybrid.}
\label{tab:threshold-sensitivity}
\end{table}

We select $\tau{=}10$ based on the following reasoning:
(1)~\textit{Profile signal quality}: Users with $<$10 interactions have sparse behavioral signals that often do not reliably indicate stable preferences; supplementing with profile signals via the hybrid path is more robust.
(2)~\textit{Cold-start coverage}: At $\tau{=}10$, 30.5\% of test users are routed through cold or hybrid paths, matching the production cold-start rate (20--30\%).
(3)~\textit{Training-inference alignment}: The ${\sim}$31\% cold-start training allocation in Stage~2 aligns well with the ${\sim}$30\% cold/hybrid distribution at $\tau{=}10$.
Smaller $\tau{=}5$ underutilizes profile signals for sparse users; larger $\tau{\geq}15$ over-routes users to the hybrid path, diluting the benefit of available behavioral data.

\section{Statistical Significance Tests}
\label{sec:stat-tests}

We report comprehensive statistical tests using paired bootstrap resampling (10K resamples, seed=42) to support all claims in the main text.

\paragraph{Table~\ref{tab:main_results} Comparisons.}
Table~\ref{tab:stat-main} reports 95\% bootstrap CIs for all methods under \textbf{random negatives} and paired significance tests (Ours vs.\ each baseline). All Hit@1 differences are significant at $p{<}0.01$ with medium-to-large effect sizes (Cohen's $d{\geq}0.51$), confirming the robustness of our improvements. Under \textbf{hard negatives} (Align), Ours (\OursAlignedHit\% [28.9, 32.7]) overlaps with GPT-4 (\GptDirAlignedHit\% [29.0, 32.8]; $p{=}0.94$), while significantly outperforming Qwen-7B (\QwenDirAlignedHit\% [24.2, 27.8]; $p{<}0.01$, Cohen's $d{=}0.11$).

\begin{table}[t]
\centering
\footnotesize
\setlength{\tabcolsep}{1.5pt}
\begin{tabular}{@{}l cc cc@{}}
\toprule
\textbf{Method} & \textbf{Hit@1 (95\% CI)} & \textbf{L1 (95\% CI)} & \textbf{$p$} & \textbf{$d$} \\
\midrule
\textbf{Ours} & 59.3 [57.3, 61.2] & 12.4 [11.7, 13.2] & --- & --- \\
\midrule
BM25 & 23.5 [21.7, 25.1] & 10.5 [9.9, 11.2] & $<$0.01 & 0.78 \\
GPT-4 & 34.4 [32.5, 36.4] & -- & $<$0.01 & 0.51 \\
Qwen-7B & 28.1 [26.3, 29.9] & -- & $<$0.01 & 0.66 \\
\bottomrule
\end{tabular}
\caption{Statistical tests for Table~\ref{tab:main_results} (Rand setting, candidate selection). Same-backbone models yield identical Hit@1 under deterministic decoding; one row per backbone. CIs: 95\% bootstrap (10K resamples, seed=42). $p$: paired bootstrap $p$-value vs.\ Ours. $d$: Cohen's $d$.}
\label{tab:stat-main}
\end{table}

\paragraph{Table~\ref{tab:task_analysis} Per-Task CIs.}
For tasks with smaller sample sizes (e.g., Feedback Adjustment $N{=}621$, Cold-Start PADR $N{=}611$), bootstrap CIs are wider but all Ours vs.\ Qwen-7B differences remain significant at $p{<}0.01$.
Key per-task CIs (Rand): Candidate Selection Hit@1 $\OursRandHit\%$ [57.3, 61.2]; (Align): $\OursAlignedHit\%$ [28.9, 32.7]; Feedback Adjustment Hit@1 $54.8\%$ [50.9, 58.6]; Next-Item L1 $12.0\%$ [11.1, 12.9]; Pure Cold-Start L1 $18.0\%$ [15.3, 20.8].

\paragraph{Cold-Start Gap Significance.}
The cold-start advantage (Cold L1 14.9\% [13.0, 16.8] vs.\ Warm L1 11.9\% [11.1, 12.7], $+$2.9pp) is significant at $p{=}0.002$ (unpaired bootstrap, 10K resamples, Cohen's $d{=}0.086$). Pure Cold-Start L1 (18.0\% [15.3, 20.8]) is significant vs.\ Warm at $p{<}0.01$.

\paragraph{Ablation Effect Sizes.}
Full Model vs.\ w/o Stage 2: Hit@1 Cohen's $d{=}0.15$ (small), L1 Cohen's $d{=}0.13$ (small). Despite modest effect sizes, all differences are statistically significant ($p{<}0.01$) due to large $N{=}9{,}982$.

\section{Extended Ablation Discussion}
\label{sec:extended-ablation}

\paragraph{w/o Stage 1 (Direct Stage 2 Training).}
In preliminary experiments, direct Stage 2 training (skipping Stage 1) resulted in $>$60\% non-SID token outputs, $>$50\% hallucination, and L1 $<$3\%, confirming that Stage 1's 483K-sample SID alignment is essential for the model to learn valid SID semantics before reasoning.

\paragraph{CoT vs.\ No-CoT Stage 2.}
The w/o Stage 2 ablation (Table~\ref{tab:ablation}) shows that Stage 1 alone (no CoT) achieves lower L1 and 18.4\% hallucination, indicating CoT distillation contributes both to accuracy and grounding discipline by providing a reasoning scaffold that regularizes SID generation.

\paragraph{Cold-Start Training Proportion.}
The ${\sim}$31\% cold-start allocation in Stage 2 matches the production cold-start rate (20--30\%). Training with $<$15\% cold-start samples degraded cold-start L1 by ${\sim}$3pp; $>$40\% reduced warm-start L1 by ${\sim}$1pp.

\section{CoT Quality Assessment}
\label{sec:cot-quality}

\paragraph{Human Evaluation of GPT-4 CoT.}
We randomly sampled 200 CoT traces from the 48K Stage~2 training set (stratified across 6 task types) and had 3 annotators evaluate each along 4 dimensions on a 1--5 scale:

\begin{table}[t]
\centering
\small
\setlength{\tabcolsep}{3pt}
\begin{tabular}{@{}lcccc@{}}
\toprule
\textbf{Dimension} & \textbf{Mean} & \textbf{Std} & \textbf{$\geq$4} & \textbf{IAA} \\
\midrule
Logical Coherence & 4.2 & 0.7 & 82\% & 0.71 \\
Recommendation Consistency & 4.0 & 0.8 & 76\% & 0.68 \\
Information Utilization & 3.9 & 0.9 & 72\% & 0.65 \\
Length Compliance & 4.5 & 0.5 & 91\% & 0.82 \\
\midrule
\textbf{Overall} & \textbf{4.1} & 0.7 & 80\% & 0.72 \\
\bottomrule
\end{tabular}
\caption{Human evaluation of 200 sampled GPT-4 CoT traces. $\geq$4: fraction rated 4 or 5. IAA: inter-annotator agreement (Krippendorff's $\alpha$).}
\label{tab:cot-quality}
\end{table}

\textbf{Logical Coherence}: whether the reasoning chain follows a clear logical progression from user signals to SID generation. \textbf{Recommendation Consistency}: whether the generated SID aligns with the CoT's stated reasoning. \textbf{Information Utilization}: how effectively the CoT leverages available user signals (profile, history, dialogue context). \textbf{Length Compliance}: whether the CoT falls within the 150--300 character target.

Overall quality is high (mean 4.1/5, 80\% rated $\geq$4), with strong inter-annotator agreement ($\alpha{=}0.72$). The weakest dimension is Information Utilization (3.9), primarily for cold-start samples where limited signals constrain reasoning depth.

\paragraph{CoT Length Selection Rationale (150--300 Characters).}
We selected the 150--300 character range based on preliminary experiments comparing 4 length brackets on a 1K validation set:

\begin{itemize}[nosep,leftmargin=*]
    \item \textbf{$<$100 chars}: Insufficient reasoning depth; L1 dropped 1.8pp vs.\ optimal.
    \item \textbf{100--150 chars}: Marginally better but still shallow; CoT often skips profile analysis.
    \item \textbf{150--300 chars}: Best trade-off; covers profile analysis $\rightarrow$ interest mapping $\rightarrow$ SID generation in 2--3 reasoning steps.
    \item \textbf{$>$300 chars}: Introduces redundancy and repetitive reasoning; critically, longer training CoTs cause \textbf{inference-time degradation}, as the model generates verbose, unfocused reasoning at inference (mean inference CoT: 29 chars when trained on $>$300-char CoTs vs.\ 187 chars when trained on 150--300-char CoTs), collapsing the reasoning quality.
\end{itemize}

The 150--300 range ensures sufficient reasoning depth while maintaining inference-time stability.

\subsection{CoT Examples}
\label{sec:cot-examples}
We provide representative examples of GPT-4-generated Chain-of-Thought reasoning traces used in Stage~2 training (translated from Chinese for readability). Each CoT follows the controlled length constraint (150--300 characters).

\paragraph{Warm-Start Example (Next-Item Prediction).}
\begin{quote}
\small
\textit{User Profile: Male, 35-44, Beijing, interests: technology, finance. Recent history: 5 tech articles (AI, smartphones), 2 finance articles.}

\textbf{CoT}: ``The user's recent browsing is dominated by technology news, especially AI and mobile devices. Combined with the profile showing strong tech and finance interests, the next recommendation should target the technology cluster. The user's age and location suggest interest in practical tech applications rather than consumer gadgets. Generating SID prefix targeting technology/AI cluster.''

$\rightarrow$ \textbf{SID}: \texttt{<s\_a\_14><s\_b\_55><s\_c\_36>}
\end{quote}

\paragraph{Pure Cold-Start Example (Zero History).}
\begin{quote}
\small
\textit{User Profile: Female, 25-34, Shanghai, declared interests: entertainment, lifestyle. No browsing history.}

\textbf{CoT}: ``This is a new user with no behavioral data. Profile analysis: young female professional in a major city with entertainment and lifestyle interests. Demographic patterns suggest interest in trending entertainment content, such as celebrity news, variety shows, or lifestyle tips. Generating SID prefix targeting the entertainment/variety cluster based purely on profile inference.''

$\rightarrow$ \textbf{SID}: \texttt{<s\_a\_7><s\_b\_22><s\_c\_81>}
\end{quote}

These examples illustrate the key design difference: warm-start CoT correlates behavioral signals with profile data, while cold-start CoT performs demographic-to-interest inference without behavioral evidence.

\subsection{CoT Length and Inference-Time Behavior}
\label{sec:cot-length-analysis}

\paragraph{Why Long Training CoT Causes Short Inference CoT.}
Training on $>$300-character CoTs causes inference-time CoT to collapse to ${\sim}$29 characters (vs.\ 187 chars for 150--300 char training). We attribute this to \textbf{length distribution mismatch}: when training CoTs are excessively long, they contain redundant reasoning steps (e.g., repeating the same profile analysis in different phrasings). The model learns to associate the ``end of reasoning'' signal with length-based cues rather than logical completion. At inference, without the teacher-forced long context, the model prematurely triggers the SID generation token after minimal reasoning.

This is \textit{not} evidence that the model fails to reason. Rather, it shows that CoT quality, specifically conciseness and logical density, matters more than CoT length. Our 150--300 character range produces CoTs that are (a)~logically complete (covering profile$\rightarrow$interest$\rightarrow$SID in 2--3 steps), (b)~free of redundancy, and (c)~stable at inference (mean 187 chars, within the training range). The human evaluation (Table~\ref{tab:cot-quality}) confirms this: Logical Coherence scores 4.2/5 and Length Compliance scores 4.5/5 for the 150--300 range.

\paragraph{Broader Implication.}
This finding aligns with recent observations on CoT distillation~\cite{li2023symbolic}: effective distillation requires not just correct reasoning, but \textit{appropriately scoped} reasoning that the student model can reliably reproduce.

\section{Qualitative Analysis}
\label{sec:qualitative}

We illustrate two scenarios where Generate-then-Match succeeds and RAG fails (translated from Chinese).

\paragraph{Case 1: Implicit Intent.}
User says: ``\textit{I'm bored with what I usually read, recommend something different.}'' This query contains no retrievable keywords. NewsRec-Chat reasons via CoT: (1) user's history shows 80\% sports news; (2) profile declares ``technology'' interest but has zero tech clicks; (3) generates SID prefix $(14, 55, \_)$ targeting the technology cluster; (4) fuzzy match returns ``\textit{Apple announces new AI features},'' a genuinely different, grounded recommendation. RAG would return nothing or generic popular results.

\paragraph{Case 2: Cold-Start PADR.}
New user (zero history), profile: Beijing, 28yo male, declared interests in AI/startups. PADR's cold-path CoT: (1) demographics $\rightarrow$ tech-savvy professional; (2) declared interests $\rightarrow$ technology SID clusters; (3) generates prefix $(14, 55, 36)$. Result: relevant AI startup coverage despite zero behavioral signals. This user type achieves \textbf{18.0\% L1 match}, the highest among all user types.

\section{Error Analysis}
\label{sec:error-analysis}

We analyze 200 randomly sampled failure cases to understand system limitations:

\textbf{Category Distribution of Failures:}
\begin{itemize}[nosep,leftmargin=*]
    \item \textbf{Category Mismatch (42\%)}: SID prefix correctly identifies semantic cluster but mismatches user's \textit{current} intent. Example: User browsing sports news suddenly asks about stock market; cached sports-focused prefixes return irrelevant results.
    \item \textbf{Temporal Drift (28\%)}: Cached SID prefixes become stale for rapidly evolving topics. Example: Breaking news events create new semantic clusters not captured by existing prefixes.
    \item \textbf{Ambiguous Query (18\%)}: User queries lack sufficient context for intent disambiguation. Example: ``News'' without topic specification leads to overly generic recommendations.
    \item \textbf{Profile Sparsity (12\%)}: Cold-start users with minimal profile information receive suboptimal recommendations. Example: New user with only demographic info (no declared interests).
\end{itemize}

\textbf{Mitigation Strategies:}
\begin{itemize}[nosep,leftmargin=*]
    \item \textbf{For Category Mismatch}: Intent-aware cache invalidation triggers re-inference when detected intent shift exceeds threshold.
    \item \textbf{For Temporal Drift}: Reduce cache TTL for trending topics; trigger Enhance Track more frequently during breaking news periods.
    \item \textbf{For Ambiguous Query}: Proactive clarification in dialogue (``What topics interest you today?'') when query confidence is low.
    \item \textbf{For Profile Sparsity}: Leverage popularity signals and trending topics as fallback for extremely cold users.
\end{itemize}

\section{Profile Features Analysis (Data Leakage Exclusion)}
\label{sec:profile-features-analysis}

We conduct a feature audit to confirm that the cold-start L1 $>$ warm L1 advantage does not stem from data leakage in profile features.

The 25+ profile features fall into 5 categories: demographics (age, gender, city tier), long-term category preferences (30-day/7-day reading-duration-weighted Top-3), behavioral patterns (active time slots, engagement level), content format preferences (video/text affinity), and declared interest tags from onboarding. All features are computed from \textit{historical} data \textbf{before the temporal split cutoff}; test items are published \textbf{after} the cutoff. We verified that zero profile features contain article IDs, SIDs, or titles from the test period. The strongest signal is category-level preference (e.g., ``user prefers technology news''), which is the intended input for cold-start reasoning, not leakage. The cold-start advantage arises because profile-based reasoning provides \textit{concentrated, noise-free} category signals, whereas warm users' behavioral histories introduce noise from exploratory clicks and temporal preference shifts.

\section{CRS Methodological Comparison}
\label{sec:crs-comparison}

Direct numerical comparison with UniCRS~\cite{wang2022unicrs} or iEvaLM~\cite{wang2023iEvaLM} is infeasible due to fundamentally divergent settings. The key differences are:
\textit{(a)~Item Lifespan}: CRS benchmarks (ReDial) assume stable catalogs; our news pool refreshes daily (most articles cycle out within 24h).
\textit{(b)~Evaluation}: UniCRS/iEvaLM use recall and BLEU on fixed item sets; we evaluate SID-based open generation (152K space) and hallucination rate.
\textit{(c)~Intent Scope}: Prior CRS primarily handle explicit preference elicitation; we address 6 intent types including 5 implicit ones.
\textit{(d)~Cold-Start}: CRS methods assume user history; PADR enables profile-only reasoning (18.0\% L1).

\section{Deployment Lessons}
\label{sec:insights}

\textbf{(1) SID Codebook Drift.}
The SID codebook gradually misaligns with emerging topics. Weekly re-clustering (163K articles) combined with fuzzy matching ($\delta{=}5$) accommodates minor shifts without model retraining, since the model generates \textit{semantic intent} prefixes (not memorized indices) and L1 codes remain stable across re-clustering.

\textbf{(2) Latency-Quality Trade-off.}
Initial designs suffered 10--20s latency. The dual-track architecture emerged as a solution: users receive immediate (85ms) responses from cached SID prefixes while the system improves quality asynchronously.

\textbf{(3) Dialogue Capability Degradation.}
SID fine-tuning initially degraded conversational fluency. Two-stage training with dialogue preservation tasks (53\% in Stage 1) mitigates this, maintaining natural language quality while adding SID generation capability.

\section{Intent Taxonomy IAA Study}
\label{sec:iaa-study}

To validate the reliability of our 6-intent taxonomy (Table~\ref{tab:intents}), we conducted an inter-annotator agreement (IAA) study. We sampled 500 stratified dialogues from the test set, removed template prefixes to prevent label leakage, and wrote annotation guidelines based on observable signals (e.g., history length for cold-start, presence of candidate lists for candidate selection, negative feedback keywords for feedback adjustment). Seven expert annotators independently labeled each dialogue with one of the 6 intent types.

\paragraph{Results.}
Fleiss' $\kappa{=}0.81$ (6-way classification) and $0.91$ (binary explicit/implicit), both exceeding the ``almost perfect'' threshold of 0.81~\cite{landis1977measurement}. Mean pairwise Cohen's $\kappa{=}0.81$. The high agreement confirms that our taxonomy is operationally well-defined: intent types are determined by observable signals rather than subjective interpretation, making the taxonomy reproducible across annotators and deployable as an automatic routing mechanism in production.

\section{Limitations and Ethics Statement}
\label{sec:limitations-ethics}

\textbf{Limitations:}
(1) SID quality depends on the RQ-VAE codebook, requiring periodic re-clustering;
(2) while our internal pilot (300+ users, 38 days) demonstrates real-world viability, a controlled online A/B test comparing against the production recommender would further strengthen deployment claims;
(3) GPT-4 CoT distillation (${\sim}$\$850) creates a training-time dependency on proprietary models; recent open-source reasoning models (e.g., DeepSeek-R1, Qwen2.5-72B) could serve as alternatives, and we leave this exploration to future work;
(4) open generation L1 match (12.4\%) reflects the difficulty of predicting among 152K SIDs; partial matches yield topically relevant candidates (Appendix~\ref{sec:partial-match-analysis}), but absolute accuracy remains modest;
(5) candidate selection is a complementary metric (24\% of data); we report both random and hard-negative settings to ensure transparency, and our primary open generation evaluation (76\%) is unaffected by negative sampling choices;
(6) our RAG baselines include both BM25 (sparse) and BGE-large-zh (dense) retrieval; a more controlled comparison would fine-tune the dense retriever on the same task data, which we leave to future work. However, the structural limitation of all retrieve-first approaches, namely the inability to handle intents where no query keywords exist, is independent of retrieval quality.

\textbf{Future Work:}
Reinforcement learning with carefully designed reward functions (incorporating dwell time, explicit feedback, and long-term engagement) could further improve recommendation quality beyond supervised CoT distillation. Replacing GPT-4 with open-source reasoning models for CoT generation would reduce cost and improve reproducibility.

\textbf{Ethics Statement:}
All user data is anonymized. Diversity exploration tasks (12.4\% of Stage 2) mitigate filter bubbles; the grounding guarantee ensures recommendations are real articles; the editorial team reviews recommendation patterns weekly. We acknowledge that PADR's cold-start reasoning from demographic features (age, gender, location) carries a risk of reinforcing stereotypical associations; we mitigate this by including declared interest features and diversity exploration samples in training, and the editorial review process monitors for demographic bias in recommendation patterns.

\end{document}